\pdfoutput=1
\documentclass[11pt,a4paper]{article}
\usepackage[hyperref]{acl2020}
\usepackage{times}
\usepackage{latexsym}
\usepackage{graphicx}

\usepackage{microtype}

\aclfinalcopy 

\title{From SPMRL to NMRL: What Did We Learn (and Unlearn) \\ in a Decade of Parsing Morphologically-Rich Languages (MRLs)?}

\author{Reut Tsarfaty  \hspace{0.1in}Dan Bareket \hspace{0.1in} Stav Klein \hspace{0.1in} Amit Seker \\
  Bar Ilan University,
  Ramat-Gan, Israel \\
  \texttt{reut.tsarfaty@biu.ac.il}\\ \texttt{\{dbareket,klein.stav,aseker00\}@gmail.com}
  }

\date{}

\begin{document}
\maketitle
\begin{abstract}
It has been exactly a decade since the first establishment of  SPMRL, a research initiative unifying  multiple research efforts to  address the peculiar challenges of Statistical Parsing for Morphologically-Rich Languages (MRLs). Here we reflect on parsing MRLs in that decade,  highlight the solutions and lessons  learned for the {\em architectural, modeling} and {\em lexical}  challenges  in the pre-neural era, and argue that similar challenges  re-emerge  in  neural architectures for MRLs. We then aim to offer a climax, suggesting that  incorporating {\em  symbolic} ideas proposed in  SPMRL terms into nowadays {\em neural} architectures has the potential to   push NLP for MRLs to a new level. We  sketch  a strategies for designing Neural Models for MRLs (NMRL), and showcase preliminary  support for these strategies via  investigating the task of {\em multi-tagging}   in  Hebrew, a morphologically-rich,  high-fusion, language.





\end{abstract}

\section{Introduction}
\label{sec:intro}
 The ability to process natural language data and to automatically extract structured meanings  out of them has always been the hallmark of Artificial Intelligence (AI), and today  it is also of immense practical value in  downstream technological applications for Information Extraction, Text Analytics, and diverse Data Science applications.
 The introduction of deep learning models \cite{deeplearn}  into Natural Language Processing (NLP)  has led to an explosion in the Neural models and pre-training techniques applied to NLP tasks --- from classical  tasks as tagging and parsing to end-to-end tasks as machine translation and question answering --- raising the performance bar on these tasks to an all-times peak.
So far though, these advances have been reported mostly for English. 
Can these advances  carry over to languages that are typologically vastly different from English, such as {\em Morphologically-Rich Languages}?

The term {\em Morphologically-Rich Languages} (MRLs) refers to languages such as  Arabic, Hebrew, Turkish or Maltese, in which significant information is expressed morphologically, e.g., via word-level variation, rather than syntactically, e.g., via fixed word-order and periphrastic constructions, as in English. 
These properties lead to  diverse and ambiguous structures, accompanied with huge lexica, which in turn make
 MRLs notoriously hard to  parse \cite{nivre07conll,tsarfaty13a}.
A decade ago, \newcite{mrls} put forth three overarching  challenges  for the MRLs research community:
\begin{itemize}
\item[] {\em (i) The Architectural Challenge:} What {\em input units} are adequate for processing MRLs? 
\\ {\em (ii)  The Modeling Challenge:} What   {\em modeling assumptions} are adequate for  MRLs?
\\{\em (iii)  The Lexical Challenge:}  How can we cope with  extreme  {\em data sparseness}  in MRLs lexica?
\end{itemize}

For  NLP  in the pre-neural era, effective solutions have been proposed and successfully applied to  address each of these challenges for MRLs,  using data from  MRLs treebanks and designated  shared tasks \cite{nivre07conll,spmrl13,spmrl14,nivre16ud}. The  solutions proposed to the above challenges  included: {\em (i)} parsing {\em morphemes} rather than words, {\em (ii)}  {\em joint modeling} of local morphology and global structures, and {\em (iii)} exploiting  {\em external knowledge} to analyze the {long tail} of un-attested word-forms.

Upon the introduction of  Neural Network models into NLP  \cite{primer}, it was hoped that we could dispense with the need to model different languages differently. Curiously though, this has not been the case. Languages with rich morphology typically require careful treatment, and often the design of additional  resources (cf.\ \newcite{tail}). Moreover, current modeling strategies for neural NLP appear to stand {\em in  contrast} with the   pre-neural proposals for processing MRLs. 

First, unsupervised pre-training techniques employing language modeling objectives (LM, MLM) are applied nowadays to  {\em raw words} rather than morphemes, and deliver {\em word-embeddings} agnostic to internal structure. While some morphological structure may be implicitly encoded in these vectors, the {\em morphemes} themselves remain un-accessible \cite{claraDEP,cotterell15}. 

Second, pre-neural  models for parsing MRLs call for {\em joint} inference over local and global structures, tasking multiple, ambiguous, morphological analyses (a.k.a.\  {\em lattices}) as input, and disambiguating these morphological structure  jointly with the parsing task   \cite{goldberg08joint,green10arabic, bohnet13joint,seeker15,more19}. In contrast, pre-trained embeddings select a single vector for each input token --- prior to any further  analysis. 

Finally,   pre-trained embeddings trained on {\em words} cannot assign vectors to unseen words. The use of unsupervised {\em char-based} or {\em sub-word} units \cite{enrich17} to remedy this situation  shows mixed results; while these models  learn {\em orthographic} similarities between seen and unseen words, they fail to learn the {\em functions} of sub-word units (\newcite{oded}; \newcite{claraLM} and references therein).

This paper aims to underscore the challenges of processing MRLs, reiterate the lessons  learned in the pre-neural era, and establish their relevance to MRL processing in neural terms. On the one hand, technical proposals as  pre-trained embeddings, fine-tuning, and end-to-end modeling,  have advanced NLP greatly. On the other hand,  neural advances often overlook MRL complexities, and disregard  strategies that were proven useful for MRLs in the past. We argue that breakthroughs in {\em Neural Models for MRLs}  (NMRL) can be obtained by incorporating  {\em symbolic}  knowledge and   {\em pre-neural} strategies   into  the end-to-end {neural} architectures. 


The remainder of this paper is organized as follows.  In Section~\ref{sec:dl} we survey the methodological changes that neural modeling brought into NLP. In Section~\ref{sec:mrl} we characterize  MRLs  and qualify the  challenges that they pose to neural NLP. In Section~\ref{sec:dlmrl} we assess the compatibility of pre-neural modeling  and  current neural modeling practices for MRLs, and in Section~\ref{sec:obj} we  suggest to re-frame pre-neural solution strategies in neural terms. In Section~\ref{sec:exp} we  present preliminary empirical support for these strategies,  and in Section~\ref{sec:con} we conclude.


\section{The Backdrop: From Classical Natural Language Processing  to End-to-End Deep Learning}
\label{sec:dl}

Classical NLP research  has been traditionally devoted to the development of  computer programs called {\em parsers}, that accept an utterance in a human language as input and  deliver its underlying linguistic structure as output. The output may  be of various sorts:
%
{\em Morphological} parsing analyzes the internal structure of  words. {\em Syntactic} parsing analyses the  structure of    sentences. 
 {\em Semantic} parsing  assigns a formal representation to the utterance, one that reflects its  meaning.   {\em Discourse} parsing    identifies the discourse units, discourse relations, as well as rhetoric and pragmatic structure associated with complete narratives.
Since natural language exhibits  ambiguity at {\em all} levels of analysis,   {\em statistical} parsers aim to  {learn} how to pick the best  analysis from multiple suitable candidates \cite{smith}. 
 The introduction of Deep Learning  has revolutionized all areas of  Artificial Intelligence, 
 and NLP research  is no exception \cite{primer}. Neural-network models now demonstrate an all-times peak in the performance of various NLP tasks, from conventional tasks in the NLP pipeline like tagging and parsing \cite{nnparsing,nntagparse, parsingall} to  diverse downstream applications,  such as  machine translation \cite{bahdanau2014neural,manning15nmt}, question answering \cite{andreas16qa}, text-to-code generation \cite{codegen} and  natural language navigation \cite{Mei}.
In addition to revolutionizing {\em empirical} NLP,   neural models  have also altered the {\em methodology} of conducting NLP research, in various ways, which we review here in turn.

First, while state-of-the-art models for  structure prediction in NLP used to rely heavily on intricate formal structures and carefully designed  features (or feature-templates) \cite{features,zhang11}, current neural models provide a form of representation learning and may be viewed as  {\em automatic feature-extractors} \cite{kiperwasser,dozat}. That is,  as long as the input object can be represented as a vector, the neural model will learn how to map it to the appropriate set of structural decisions, without having to write features or feature-templates by hand.

Second, most neural models for NLP  rely on {\em pre-training},  the process of acquiring word-level vector representations termed  {\em word-embeddings}. These vectors are used as input, instead of  actual words.
Initially, word embeddings were  {\em non-contextualized} \cite{w2v,glove}, i.e., they assigned the same vector to the occurrences of a word  in different contexts.  Later models present  {\em contextualized} embeddings \cite{bert,elmo,xlnet,roberta},  they  assign different vectors   to the occurrences of the same word in different contexts. 
Embeddings in general, and contextualized ones   in particular,  dramatically increased the performance of any  NLP task they were applied to.

Third, working with contextualized embeddings has been so successful, that it shifted the focus of  NLP practitioners  from training models from scratch to {\em fine-tuning} \cite{inoculation} pre-trained embeddings. That is, instead of tailoring hugely complex models for  specific tasks and training them from scratch, a huge effort is invested in learning a general language  model (LM) that can assign contextualized embeddings to words. These vectors are often argued to capture, or {\em encode}, various aspects of   structure and meaning \cite{hewitt19}, and then, a relatively small amount of task-specific data may be used to {fine-tune} the pre-trained embeddings, so that the model can solve a particular task at hand. 


Finally, traditional NLP  tasks, such as the parsing layers mentioned earlier, were typically organized into a pipeline turning unstructured texts gradually into more complex structures by gradually increasing the complexity of analysis. Eventually, complex semantic structures formed the basis for the design of dialogue systems, question answering systems, etc. Nowadays, NN models for  complex semantic tasks are often designed and trained {\em end-to-end (E2E)} on examples of input-output pairs. There is an implicit assumption that all relavnt linguistic features are already {\em encoded} in the pre-trained representations, and that they will be {\em automatically extracted} in the learning process.  

This methodology of  {\em pre-training}, {\em automatic feature extraction} and {\em fine-tuning} has been  applied to a wide variety of  tasks and saw immense success for English --- and also for similar languages. 
Notwithstanding, the  majority of  achievements and results for complex {\em natural language understanding} (NLU)  does not yet carry over to all languages, and in particular,  for languages known as  {\em Morphologically-Rich Languages}.


 \section{The Challenge: NLP for Morphologically-Rich Languages}
 \label{sec:mrl}
 
The term {\em Morphologically-Rich Languages} entered the NLP research community about a decade ago \cite{mrls} bringing to the forefront of the research a set of languages which are typologically  different from English and share a host of similar processing challenges. 
Subsequent SPMRL events and shared tasks \cite{seddah13,tsarfaty13a,seddah14}  illustrated how methodologies and modeling assumptions  for English NLP  often break down in the face of such typologically diversity. That is, while  most NLP models  can {\em in principle} be trained on data in any given language,\footnote{E.g., via applying them to the {universal dependencies} (UD) treebanks \cite{nivre16ud}.} such models are often developed with English in mind, and the {\em  bias} injected into such models  is not optimal for languages that  exhibit flexible word order, and rich word-internal structure, as is the case in MRLs.

Let us  briefly survey the properties of MRLs and the challenges associated with them, 
and observe how pre-neural studies  proposed to address them.
%

 \paragraph{The Essence of MRLs.}

The term {\em morphologically rich languages} (MRLs) refers to languages in which significant information regarding the  units in the sentence and  the relations between them  is expressed morphologically, i.e., via word structure, rather than syntactically, e.g., using  word order and rigid structures.
Morphologically-marked information may be of various sorts. For example, consider the following Hebrew sentence:\footnote{In the transliteration  of Simaan et al.\ \cite{simaan01treebank}.} 
\begin{itemize}
\item[(1)]  {\em hild hpil at hspr fl hildh.}\\
literally: the-kid.MASC.SING cause-to-fall.MASC.PAST ACC the-book of the-kid.FEM.SING\\
trans: ``the boy made the book of the girl fall.''
\end{itemize}
There are several lessons to be learned from (1).  First note that the 6 tokens  in Hebrew correspond to 9 tokens  in the English translation --- we can observe three types of morphological phenomena that has led to this. First, elements such as  prepositions, relativizers and the definite markers  {\em h} (the) in Hebrew always attach as {\sc clitics} to lexical hosts, and do not stand on their own. Second, features  as gender, number, person, tense etc.\ are marked by {\sc inflectional} morphemes. In particular,   the final {\em h} distinguishes {\em ildh} kid.FEM  from  its  {\em ild} kid.MASC counterpart. Interestingly, an  initial {\em h}   marks definiteness in {\em hild, hspr} and {\em hildh}, so there is no 1:1 relation between surface elements (chars) and what they can mark. Finally, the Hebrew verb, {\em hpil}, which also begins with an {\em h},  corresponds to the construction (``{\em binyan}'', pattern) `cause-to-fall' via a {\sc derivational} morphological process that combines the pattern {\em h\(\_\_\)i\(\_\)}  (causative) and the lexical root  {\em n.p.l} (to fall). Note that  the {\em h\(\_\_\)i\(\_\)}  causative morpheme  is non-concatenative. Moreover, when
 combining {\em h\(\_\_\)i\(\_\)} + {\em n.p.l} into {\em hpil} the {\em n} drops, leaving  only a part of the root explicit.  

This word-level complexity then requires decomposition of raw surface tokens into constituent morphemes in order  to transfer them to the syntactic, semantic, or downstream tasks that require this information. However, rich morphology may  lead to extreme ambiguity in the decomposition of   tokens into morphemes. 
Take for example the two occurrences of the  word form {\em hpil} in (2):
\begin{itemize}
\item[(2)]  {\em hild hpil at hpil.}\\
literally: the-kid.MASC.SING cause-to-fall.MASC.PAST ACC the-elephant\\
translated: ``the boy made the elephant fall.''
\end{itemize}
Two different morphological processes lead to two different  decompositions of {\em hpil}, one is concatenative: ``the'' +  ``elephant" ({\em h+pil}) and one is not: ``cause-to'' + ``fall'' ({\em h\(\_\_\)i\(\_\)} + n.p.l). Moreover, neither interpretation is a-priory more likely than the other. We need the global context in order to select the single human-perceived analysis for each form.

 \paragraph*{The Typology of MRLs.}
The extent to which morphological phenomena is reflected in different languages varies, and 
linguistic typology describes  morphological diversity along two dimensions. 
One is the {\em synthesis} dimension, which captures the ratio of {\em morphemes per word}. {\em Isolating} languages on one end present one-morpheme-per-word, like most words in English. At the other end  we have {\em polysynthetic languages}, where multiple morphemes can form  a single  word,  as it is in Turkish. 
The other dimension is  {\em fusion}, and it refers to {\em how easy} it is to decompose the word into morphemes. In Turkish, which is {\em agglutinative}, the segmentation   into morphemes is rather straight-forward. This  stands in contrast with {\em fusional} languages, such as Hebrew, where the decomposition of a word like {\em hpil}  is less trivial due to the intricate `fusion' processes that went into  creation. 


 \paragraph*{Key Challenges in NLP for MRLs}
 
 The  linguistic characteristics of MRLs are known to pose  challenges to the development of NLP models,   shared across languages and tasks. The overarching challenges are summerized in  \newcite{mrls}:
\begin{itemize}
\item[] (i) {\sc The Architectural Challenge}: What are the units that should enter as input into the NLP pipeline for MRLs? Are they words? Morphemes? How are these units identified and propagated down the pipeline?
\\ (ii) {\sc The modeling Challenge}:  What are the modeling assumptions that are appropriate for models for MRLs? What kind of structure representations and  features (or feature-templates) are appropriate?
\\(iii)  {\sc The Lexical Challenge}:  How can we cope with the extreme data sparseness that follows from the  complex structure of words and the productivity of  morphology?
\end{itemize}

 \paragraph*{Pre-Neural Solutions in NLP for MRLs.}
Let us now survey the solutions proposed for these three overarching challenges in the {\em pre-neural} era.

In response to the {\sc architectural} challenge, several input alternatives have been proposed. The input to processing an MRL can be composed of raw tokens, segmented morphemes, or complete morphological lattices that capture the multiple possible analyses for each input tokens \cite{amir-conll}. Morphological lattices seem particularly advantageous, since on the one hand they represent the explicit decomposition of  words into morphemes, and on the other hand retain the morphological ambiguity of the input stream, to be  disambiguated downstream, when information from later phases, syntactic or semantic, becomes available. 

Lattice-based processing has led to re-thinking the {\sc modeling} architectures for MRLs, and to propose {\sc joint} models, where multiple levels of information are represented during training, and  are jointly predicted at inference time. 
Such joint models have been developed  for MRLs in the context of phrase-structure parsing  \cite{tsarfaty06,goldberg08joint,green10arabic} and   dependency parsing \cite{bohnet,centinoglu,more19}. In all cases, it has been shown that  {\em joint} models obtain better results than their morphological or syntactic standalone counterparts.\footnote{Joint models are shown to be effective  for other tasks and languages, such as  parsing and NER \cite{jenny} or parsing  and SRL \cite{depsrl}.}

Finally, the {\sc lexical} challenge  refers to the problem of {\em out-of-vocabulary}  items. Supervised training successfully  analyzes attested forms, but fails to analyze the long tail of  morphological forms  in the language,  not yet attested during training. Pre-neural models for  MRLs thus benefit from additional {\em symbolic} information  beyond the supervised data. It can be in the form of online dictionaries, wide-coverage lexica,  or a-priori knowledge of the structure of  morphological paradigms in the language \cite{lefff,meni09enhance}.

\paragraph{Where We're At}
Upon the introduction of   neural models into NLP the hope was that  
we could dispense with the need to develop language-specific  modeling strategies, and that  models  will seamlessly carry over from any one language (type) to another. Curiously, this was not yet shown to be the case. NLP advances in MRLs  still lag behind those  for English, with lower empirical results on classical tasks  \cite{udpipe}, and very scarce results for applications  as question answering and natural language inference \cite{xtreme}.

More fundamentally, NLP researchers nowadays successfully {predict}  linguistic properties of English via neural models as in \newcite{linzen,gulordava}, but they are  less successful in doing so for  languages that  differ from English, as in \newcite{shauli}.  It is high time for the MRL community to  shed light on the  methodological and empirical gaps  between neural models for English and for MRLs, and to bridge this gap.

 \section{The Research Objective: NLP for MRLs in the Deep Learning Era} 
\label{sec:dlmrl}

%
The point of departure of this  paper is the claim that neural modeling practices employed in NLP nowadays  are {\em suboptimal} in the face of properties of MRLs. 
In what follows we illuminate this claim for the four neural methodological constructs that we termed {\em  pre-training}, {\em  fine-tuning}, {\em   feature-extraction} and {\em   end-to-end modeling}.

{\em Pre-training} of word embeddings presupposes that the input to an NLP architecture consists of raw words. 
 However,  {word}-level embeddings  may not be useful for  tasks that require access to  the actual morphemes. For example,   for  semantic tasks in MRLs, it is  often better to use {\em morphological embeddings} of lemmas rather than words \cite{oded}. Also, dependency parsing for MRLs requires access to morphological segments, according to the UD scheme \cite{udpipe}. 
 
A reasonable solution might  be to morphologically analyze and segment all input words  {\em prior to} pre-training. Unfortunately, this solution does not fit the bill for MRLs either. First, current neural segmentors and taggers for MRLs are not accurate enough, and errors in the analyses propagate through the pre-training to contaminate the  trained embeddings and later tasks. In the universal segmentation work of \cite{shao18seg}, for instance, neural segmentation for  languages which are high on both the {\em synthesis} and the {\em fusion} index, such as Arabic and Hebrew, lags far behind. 
Beyond that, there is the technical matter of resources. Pre-training  models  as  \newcite{bert,roberta,xlnet}  requires  massive amounts of  data and computing resources, and such training often takes place outside of academia. 
Training {\em morphological} embeddings rather than {\em word} embeddings was not taken up by any commercial  partner.\footnote{Possibly since  this does not align with the business goals.} 

Next, let us turn to the notion of {\em fine-tuning}, widely used today in all  sorts of NLP tasks, typically in conjunction with contextualized  embeddings  as  \cite{bert,elmo,roberta}. An argument may be advanced that  {\em  contextualized  embeddings}   actually  encode accurate disambiguated morphological analyses in their context-based representations, and all we have to do is to {\em probe} these vectors and make these morphological analyses explicit. This argument is appealing, but it was never seriously tested empirically, and it is an open question whether we can successfully probe the fine-grained morphological functions from these vectors.  

A possible caveat for this line of research has to do with the inner-working of  contextualized representations. Most contextualized embeddings  operate not on words but  on {\em word-pieces}. A word-pieces algorithm   breaks down  words into  sub-words, and the model assigns vectors to  them. The word-pieces representations are later {\em concatenated} or pooled together to represent complete words. It is an open  question  
 whether these word-pieces capture  relevant aspects of morphology. In particular, it is unclear that the strategy of relying on chars or char-strings is adequate for encoding  {\em non-concatenative} phenomena that go beyond simple character sequencing, such as  templatic morphology, substraction, reduplication, and more  \cite{morphology,wp}. 

The notion of word-pieces leads us  to   consider the {\sc lexical} challenge. The suggestion to use  sub-word units (chars or char n-grams) rather than words  could naturally help in generalizing from seen to unseen word tokens.
There is  a range of sub-word units that are currently employed (chars, char-grams, BPEs \cite{rico}),  nicely compared and contrasted by \newcite{claraLM}.   \newcite{claraLM,claraDEP}
 show that
 for the type of sub-word units that are currently used,  standard  pre-training  leads to clustering  words that are similar {\em orthographically}, and do not necessarily share their linguistic {\em functions}. When a downstream task  requires   the morphological signature  (e.g.,  dependency parsing in \cite{claraDEP}) this information is not recoverable from  models based on  sub-word units alone. 

On the whole, it seems that {\em end-to-end  modeling}  for MRLs cannot completely rely on {\em automatic feature extraction} and 
dispense with the need to explicitly model morphology. It is rather the contrary.   Explicit morphological analyses  provide an {excellent} basis for successful  feature extraction and accurate downstream tasks. When such analysis is missing, results for MRLs deteriorate. So, we should aim to recover  morphological structures rather than ignore them, or jointly infer such information together with the downstream tasks.\footnote{Furthermore,  \newcite{gonen} have  recently shown that one needs to know the {\em explicit} morphological analyses in order to effectively  {\em ignore} or neutralize certain morphemes, for instance discarding gender for reducing bias in the data.}

A different, however related,  note concerning {\em automatic feature extraction} in MRLs has to do with the flexible or free word-order patterns that are exhibited by many MRLs. Many neural models  rely on RNNs \cite{lstm}   for feature extraction. These models assume complete linear ordering of the words and heavily rely on positions in the process of representation learning. Even pre-training based on {\em attention} and {\em self-attention}  \cite{allyouneed} 
 assign weights to  {\em positional} embeddings.
In this sense, the  bias of current neural models to encode {\em positions} stands in contrast with the properties of  MRLs, that often show  discrepancies between the linear position of words and their linguistic functions. It is an open question whether there are more adequate architectures for training (or pre-training) for more {\em flexible} or {\em free word-order} languages.


\section{{{Research Questions and Strategies}}}\label{sec:obj}

\paragraph*{The Overarching Goal}
The purpose of the proposed research theme, which we henceforth refer to as {\em Neural Models for MRLs} (NMRL), is to devise modeling strategies for MRLs,
for classical NLP tasks   (tagging, parsing)  and for downstream  language understanding tasks (question answering, information extraction, NL inference, and more).  This  research diverges  from the standard  methodology of applying DL for  NLP  in   three  ways.

First,  current {\em end-to-end} neural models for  complex language  understanding  are developed  mostly for English \cite{glue,superglue}. Here we aim to  situate neural modeling of natural language understanding in cross-linguistic settings (e.g., \cite{xtreme}).
Second, while current neural models for NLP assume {pre-training} with massive amounts  of unsupervised data \cite{mlbert,xlnet,roberta},  research on MRLs  might be realistically faced  with resource-scarce settings, and will require models that are more ``green''  \cite{green}.
Finally, while many neural-based models developed for English   presuppose that  linguistic  information relevant for the downstream task is implicitly encoded in word vectors, and may be successfully predicted  by neural models \cite{linzen},  we  {\em question} the assumption that ready-made pre-trained embeddings,  will indeed  encode all  relevant  information required for end-to-end models in MRLs.

The key strategies we propose in order to address NMRL include transitionining to (i) {\em morphological-embeddings}, (ii) {\em joint lattice-based modeling},  and (iii) {\em paradigm cell-filling}  \cite{wp,analogy}, as we detail shortly. 
 

\paragraph*{Research Questions.}

To instigate  research on NMRL, let us define the three overarching {\sc Deep} challenges of MRLs in the spirit of \cite{mrls}.
For these challenges, the aim is to devise solutions that respect the linguistic complexities  while employing the most recent  deep learning advances.

\begin{itemize}
\item {\sc The {Deep}  Architectural Challenge}:  The `classical' architectural challenge aimed to define optimal input and output units adequate for processing MRLs. In neural terms, this challenge   boils down to  a question concerning 
the {\em units} that should enter pre-training. Are they words? Word-pieces? Segmented morphemes?  Lemmas?  Lattices? Furthermore, should  these units be predicted from existing pre-trained embeddings (e.g., multilingual BERT \cite{mlbert} or XLNet \cite{xlnet}), or should we develop new pre-training paradigms that will make the relevant morphological units more explicit?
\item {\sc The {Deep} Modeling Challenge}:  The use of neural models for NLP tasks  re-opens an old debate concerning  {\em joint} vs {\em pipeline} architectures for parsing MRLs. The strategy of {\em pre-training} word vectors and then employing {\em feature extraction} or {\em fine-tuning}  pre-supposes a pipeline  architecture, where a model sets all morphological decisions during {\em pre-training}. Joint models assume lattices that encode ambiguity and partial order, and morphological disambiguation  happens only later, in the global context of the task. Is it possible to devise  neural  {\em joint models} parsing for MRLs? And if so, would they still outperform  a pipeline?
\item {\sc The {Deep}  Lexical Challenge}:  Despite the  reliance on pre-trained embeddings and unsupervised data, there is still an extreme amount of unseen lexical items in the long tail of inflected forms in the language,  due to the productive nature of morphology.
Therefore, we need to effectively handle words  outside of the {\em pre-trained} vocabulary. How can we cope with the extreme data sparseness in highly synthetic  MRLs? Should we incorporate external resources  --- such as dictionaries, lexica, or  knowledge of paradigm structure   --- and if so, how should such symbolic information be incorporated into the end-to-end neural model? 
\end{itemize}

\paragraph*{Solution Strategies.}

The work on NMRL may proceed along either of these four reserch avenues, each of which groups together research efforts to address a different challenge of NMRL. 


\begin{itemize} 
\item {\bf Neural Language Modeling for MRLs.}
The strategy here is  to empirically {examine} the ability of existing  pre-trained language models to encode rich word-internal   structures, and to  devise new  alternatives for {pre-training}  that would inject relevant biases  into the language models, and make morphological information effectively learnable. This may be done by  proposing better {\em word-pieces} algorithms,   and/or devising new pre-training objectives (e.g., lattice-based) that are more appropriate for MRLs.
\item {\bf Joint Neural  Models for MRLs.}
The aim here is to devise neural models that parse  morphologically ambiguous input words in conjunction to analyzing deeper linguistic  layers, and to investigate whether these joint models work better than a pipeline -- as has been  the case in pre-neural models. Neural  modeling of morphology may be donw jointly with,  named-entity recognition,  syntactic or semantic parsing, and downstream tasks as information extraction and question answering. Interleving information from all layers may be done by all at once (e.g., via Multi-Task Learning \cite{mtl}) or by  gradually adding complexity (e.g., via  Curriculum Learning \cite{curri09}). 
\item {\bf Neural Applications for MRLs.}
We aim to develop  effective strategies for devising {\em end-to-end} models for  complex language understanding  in MRLs.
To do so, the community needs high-quality benchmarks for question answering, machine reading and  machine reasoning for MRLs. Initially, we need to rely on lessons  learned concerning pre-training and joint modeling in the previous items, in order to devise successful architectures for solving these tasks.
Moreover, developing  benchmarks  and annotating them both at the morphological level and for the downstream task will help to evaluate  the benefits of explicit morphological modeling versus representation learning, for acquiring word-internal information needed for the downstream task. 
\item {\bf Closing the Lexical Gap for MRLs.}
Finally, we need to develop effective strategies for handling  out-of-vocabulary (OOV) items in neural models for MRLs. Currently, the main focus of investigation lies in breaking  words into  pieces, to help generalize from seen to unseen word tokens. As a complementary area of investigation,  a plausible direction would be to shift the focus from the {\em decomposition} of  words into morphemes, to  the  {\em organization} of words as complete  paradigms. That is, instead of  relying  on sub-word units, identify  sets of words organized into morphological paradigms \cite{wp}. Rather than construct new words from observed pieces,   complete unseen  {\em paradigms}  by analogy based on  observed complete paradigms. 
\end{itemize}

\begin{table*}[t]
\centering
\scalebox{0.85}{
\begin{tabular}{|l||c||cccc|c|c|}
\hline
Model \(\rightarrow\) & Pre-Neural & LSTM-CRF & LSTM-CRF & LSTM-CRF & LSTM-CRF  & Seq2Seq & BERT\\
  & SOTA &  & +Char & +FT & +Char+FT  &  COPYNET & \\
Segmentation \(\downarrow\) &  & & & & & &
 \\\hline
  \hline
 {\em Oracle}  & - & 91.46 &	93.2	 & 94.6	& 96.03  & - & 95.56 \\
 \hline
{\em Predicted} & - & 86.16	&86.57	&90.76	&92.57 & - & 92.27\\
{\em Raw Tokens} & - & 73.38	& 79.26	& 88.63	& 91.81 & - & 92.57 \\
{\em Raw Lattices} & {\bf 95.5} & NA	& NA & 	NA & NA	& {\bf 95.1} &  NA \\

\hline\hline
\end{tabular}}
\caption{F-Scores for Hebrew Multi-tagging  on the standard dev-set of the Modern Hebrew treebank. +Char means a character-based LSTM encoding, +FT means morphologically-trained Fast-text embeddings. BERT means fine-tuning the contextualized embeddings of Multilingual BERT \cite{mlbert}.  COPYNET is the model we propose in Section~\ref{sec:exp}. {\em Oracle} Segmentation means that the segmentation into morphemes  (expert annotation) is known in advance. {\em Predicted} Segmentation means the decomposition into morpheme   automatically predicted via  \newcite{more19}. {\em Raw Tokens} means that raw input tokens are provided as is,  {\em Raw Lattices} means that the  tokens are automatically transformed into complete  morphological lattices based on a wide-coverage symbolic lexicon.  
}\label{tab:exp}
\end{table*}

\paragraph*{{{Expected Significance.}}}

As has been the case with  SPMRL, work on NMRL
is expected to deliver architectures and  modeling strategies  that can carry across MRLs, along with a family of algorithms for predicting, and benchmarks for evaluating,  a range of linguistic phenomena in MRLs.
From a {\em scientific} standpoint, this investigation will advance our understanding of what types of linguistic phenomena neural models can encode, and in what ways properties of the language should guide the  choice of our neural architectural decisions.
 From a {\em technological} point of view, such modeling strategies will have vast applications in serving language technology and artificial intelligence advances to a range of  languages which  do not currently enjoy these technological benefits.
 %


\section{Preliminary Empirical Evidence}
\label{sec:exp}
\paragraph{Goal.}
In this section we aim to empirically assess the ability of neural models to recover the word-internal structure of  morphologically complex and highly ambiguous surface tokens in  Modern Hebrew. Hebrew is a Semitic language which lies high on both the {\em synthesis} and {\em fusion} typological indices, and thus provides an interesting case study. 

Specifically, we devised a {\em multi-tagging}  task where  each raw input token is tagged with the sequence of Part-of-Speech tags that represent the  functions of its constituent morphemes. For example, the  token    {\em hpil} in Section~\ref{sec:mrl}  can assume two different {\em multi-tag} analyses: VERB (made-fall) or  DET+NOUN (the elephant). The number of distinct tags in the {\em multi-tagging} analyses of Hebrew tokens can be up to seven different tags, that represent distinct functions contained in the word token. 

\paragraph{Models.}
We compare the results of multi-tagging obtained by  a state-of-the-art, pre-neural, morpho-syntactic parser \cite{more19} that is   based on the structured prediction framework of \newcite{zhang11framework}. 

The pre-neural parser explicitly incorporates three  components   for addressing the challenges associated with MRLs: (i) it receives complete {\em morphological  lattices} as input, where each input token is initially assigned the set of all possible morphological analyses for this token, according to a wide-coverage lexicon, (ii) it employs {\em joint training and inference} of morphological segmentation and syntactic dependencies, and (iii) it employs   {unknown-words heuristics} based on {\em linguistic rules} to assert  possible valid analyses of  OOV tokens. 

We compared the performance of this pre-neural parser to three neural architectures: 
\begin{itemize}
    \item An
end-to-end language-agnostic {\em LSTM-CRF} architecture, trained to predict a single complex tag (multi-tag) per token, encoding words with and without {\em morph/char embeddings}.
\item An architecture based on the Hebrew section  of  {\em multilingual BERT}, fine-tuned to predict a single complex tag (multi-tag) per token. 
\item As a first approximation of incorporating symbolic  morphological constructs into the neural end-to-end architecture, we  designed  our own COPYNET, a sequence-to-sequence pointer-network where the input consists of   {\em complete  morphological lattices} for each token, and a {\em copy-attention} mechanism  is trained to jointly select morphological segments and tag associations {\em from within the lattice}, to construct  the complete multi-tag analyses. 
\end{itemize}

\paragraph{Data and Metrics.}
We use the Hebrew section of the SPMRL shared task \cite{seddah13} using the standard split, training on 5000 sentences and evaluating on 500 sentences. 
For generating the lattices we rely on a rule-based algorithm we devised on top of the wide-coverage lexicon of \cite{adler06}, the same lexicon employed in previous work on Hebrew \cite{more16coling,more19,hebnlp}. We report the  F-Scores  on Seg/POS as defined in \newcite{more16coling,more19}. 

\paragraph{Results.}


Table \ref{tab:exp} shows the multi-tagging results for the different  models. The pre-neural  model 
obtains 95.5 F1 on joint Seg+POS prediction on the standard dev set. As for the neural models, in an {\em oracle}  segmentation scenario,  where the gold morphological segmentation is known in advance, both BERT and the LSTM-CRF get close to the pre-neural model results. However,  {they solve an easier and {\em unrealistic} task}, since in realistic scenarios the gold segmentation is never known  in advance. 
In the  more {\em realistic} scenarios, where the segmentation is automatically predicted  (via  \newcite{more19}), the results   of the Neural models substantially drop. 
As expected, morph-based and char-based representations help to improve results of the LSTM-CRF model, though not yet  reaching the 95 F-score of the pre-neural model.
Finally, employing our COPYNET with symbolic morphological lattices,  with  OOV  segmentation heuristics as in the pre-neural model, leads to the most significant improvement, almost  closing the gap with the pre-neural state-of-the-art result. Unfortunately, lattices are {\em  incompatible}  with LSTMs and with {BERT}, since LSTMs and BERT models  assume complete linear ordering of the tokens, while lattices impose only a {\em partial order} on the morphemes. The question how to incorporate contextualized embeddings into joint, lattice-based,  models is fascinating, and calls for  further research.

\section{Discussion and Conclusion}
This  paper proposes NMRL, a new (or rather, redefined) research theme  aiming to develop neural models, benchmarks, and modeling strategies for MRLs. We surveyed  current research practices in neural  NLP,  characterized the particular  challenges associated with  MRLs, and  demonstrated that some of the  neural modeling practices are incompatible with the accumulated wisdom concerning  MRLs in the SPMRL literature.

We proceeded to define the three {\sc deep}  counterparts to the challenges proposed in \newcite{mrls}, namely, the {\sc Deep Architectural Challenge, Deep Modeling Challenge} and {\sc Deep Lexical Challenge}, and sketched  plausible research avenues  that the NMRL community might wish to explore towards their resolution. 

Our preliminary experiments on Hebrew multi-tagging confirmed that relying on lessons learned for MRLs in the pre-neural era and incorporating similar theoretical constructs into the neural architecture indeed  improves the empirical results on {\em multi-tagging} of  Hebrew, on the very basic form of analysis of Modern Hebrew ---  a morphologically rich and highly-fusional language. 

This type of research needs to be  extended to the investigation of multiple tasks, multiple languages, and multiple possible pre-training regimes (words, chars, morphemes, lattices)
in order to investigate whether this trend extends to other languages and tasks.
Whether adopting   solution strategies for MRLs proposed herein or devising new ones,  it is high time to bring the linguistic   and moprhological complexity of MRLs back to the forefront of NLP research, both for the purpose of getting a better grasp of the abilities,  as well as limitations, of  neural models for NLP, and  towards serving the exciting NLP/AI advances to the understudied,  less-privileged, languages. 
\label{sec:con}

\section*{Acknowledgments}
We thank Clara Vania, Adam Lopez, and members of the Edinburgh-NLP seminar,  
 Yoav Goldberg, Ido Dagan, and members of the BIU-NLP seminar, for intriguing discussions on earlier presentations of this work.
This research is kindly supported by the Israel Science Foundation (ISF), grant No.\ 1739/16, and by  the  European Research Council (ERC),  under  the  Europoean Union Horizon 2020 research and innovation programme,  grant  No.\ 677352.

\bibliographystyle{acl_natbib}  
\bibliography{acl2020.bib}

\end{document}